# Integrated Framework for Selecting and Enhancing Ancient Marathi Inscription Images from Stone, Metal Plate, and Paper Documents

**Bapu D. Chendage[1], Rajivkumar S. Mente[2]**

[1,2] Department of Computer Applications, P.A. H. Solapur University, Solapur, Maharashtra,
Email Id: bdchendage@sus.ac.in, rsmente@sus.ac.in

**Abstract**

Ancient script images often suffer from severe background noise, low contrast, and degradation caused by aging and environmental effects. In many cases, the foreground text and background exhibit similar visual characteristics, making the inscriptions difficult to read. The primary objective of image enhancement is to improve the readability of such degraded ancient images. This paper presents an image enhancement approach based on binarization and complementary preprocessing techniques for removing stains and enhancing unclear ancient text. The proposed methods are evaluated on different types of ancient scripts, including inscriptions on stone, metal plates, and historical documents. Experimental results show that the proposed approach achieves classification accuracies of 55.7%, 62%, and 65.6% for stone, metal plate, and document scripts, respectively, using the K-Nearest Neighbor (K-NN) classifier. Using the Support Vector Machine (SVM) classifier, accuracies of 53.2%, 59.5%, and 67.8% are obtained. The results demonstrate the effectiveness of the proposed enhancement method in improving the readability of ancient Marathi inscription images.

**Keywords**

Ancient Marathi Script, Inscription Image Processing, Background Noise Reduction, Stain Removal, Binarization Techniques, KNN Classifier, SVM Classifier

**Introduction**

The images of ancient scripts have great value in epigraphy as they form the primary source of historical, cultural, and linguistic heritage. Epigraphy as an area of research involves the study of inscriptions carved on stones, metal plates, temple walls, and pillars, among others [1], [2]. But the processing and analysis of these inscriptions pose a problem as they get severely deteriorated due to their prolonged stay in the atmosphere of rain, wind, sun, and extreme temperatures, resulting in images of low contrast, poor resolution, and dim characters. Additionally, the presence of the inscription at different locations in the surroundings does not allow the image acquisition to have proper resolution due to irregularities in the textures.

Traditionally, paper squeezes, rubbing, and scale drawings have been adopted to document inscriptions on paper or paper-based materials to record the text contents. Nonetheless, these approaches require human labor and are associated with loss of detail and addressing complex degradation issues within digital copies of inscriptions. Recently, in the study of document and ancient inscriptions preprocessing, diverse state-of-the-art approaches have been put forward to improve document readability to ease automations. Some of the approaches include document analysis and processing associated with noise removal, contrast enhancement, and morphological processing in document readability before binarization [1],[2], and recent approaches employing deep leaning for adaptive improvement of document enhancement and binarization through vision transformer and generative models [3],[4].

One of the most important preprocessing tasks in ancient script analysis is binarization, which is concerned with highlighting the foreground text by segregating it from all surroundings. Binarization involves converting a grayscale or color image to a binary form in which pixels are identified as either foreground (text) or background [5]. Thresholding is one of the most prominent aspects of binarization in which images are divided into global or local segmentation techniques. Otsu's global segmentation algorithm is ideal for images with consistent lighting conditions, whereas images with high variations in light and noise in the background can be efficiently binarized using local adapted thresholding techniques in which threshold values are determined through neighborhood statistics such as mean and variance values.

Although the classical thresholding method is efficient, ancient images may need combined preprocessing and adaptive thresholding techniques. Some recent research has investigated combined noise removal and adaptive thresholding approaches, as well as learning schemes that model both global and local information using image enhancement techniques for binarization tasks [6],[7], which refers to current trends and challenges in document image binarization. In the presented work, an image preprocessing technique using binarization and complementary techniques has been proposed. The performance of the proposed techniques has been tested on the stone, metal plate, and document script datasets. The improvement in the readability of the ancient script images has been measured using the performance of the classifier.

**Local Enhancement Using Mean and Standard Deviation**

Mean and standard deviation are fundamental statistical measures widely used in image processing for local enhancement. The mean represents the average intensity of pixels in a local neighborhood and provides an estimate of the overall illumination in a grayscale image region. Standard deviation quantifies the spread of intensity values around the mean, indicating the variability within the local region. A low standard deviation implies that pixel values are clustered closely around the mean, whereas a high standard deviation indicates significant intensity variations, which are often associated with edges, textures, or noise [8],[9],[10].

For a local neighborhood containing n pixel intensity values xi, the mean μ is calculated as:

$$\mu = \frac{1}{n}\sum_{i=1}^{n} x_i \tag{1}$$

The standard deviation σ\sigmaσ is calculated as:

$$\sigma = \sqrt{\frac{1}{n}\sum_{i=1}^{n}(x_i - \mu)} \tag{2}$$

These local statistics are utilized in adaptive image enhancement to improve contrast and to effectively separate foreground text from background noise in ancient Marathi script images [11], [15],[17].

## Methodology

The proposed methodology aims to enhance ancient Marathi script images by reducing noise and improving overall image quality. The preprocessing begins with noise reduction and quality enhancement, followed by extraction of text regions using morphological operations. For each extracted text region, local statistical measures mean and standard deviations are computed to characterize texture and illumination variations. Images with irregular or degraded backgrounds typically exhibit higher standard deviation values compared to those with uniform backgrounds [12] [13].

The enhanced images are then classified using K-Nearest Neighbor (K-NN) and Support Vector Machine (SVM) classifiers. The dataset used in this study comprises 25 images, including 10 ancient Marathi script images engraved on stone, 10 on metal plates, and 5 document-type images [14],[16]. The images are categorized into regular background (normal) and irregular background (abnormal) types for analysis. Experimental results indicate that document-type images achieve higher classification accuracy compared to scripts engraved on stone and metal plates, highlighting the effectiveness of the proposed enhancement approach for certain image types.

The processing stages of the proposed methodology are illustrated in Figures 1–3. Figure 1 shows the stages applied to script images engraved on stone, Figure 2 demonstrates the methodology for document-type images, and Figure 3 presents the workflow for script images on metal plates.

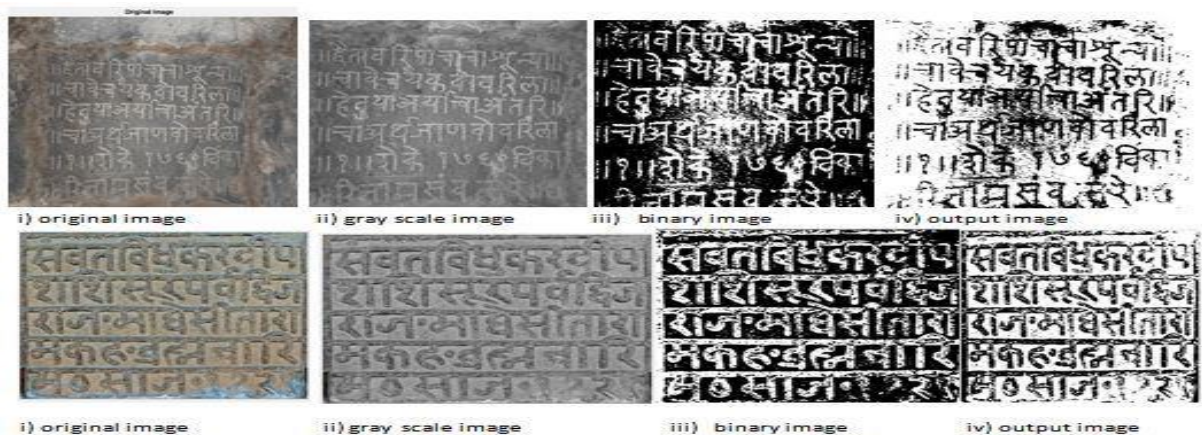

**Fig.1 Stages of the proposed method on stone images**

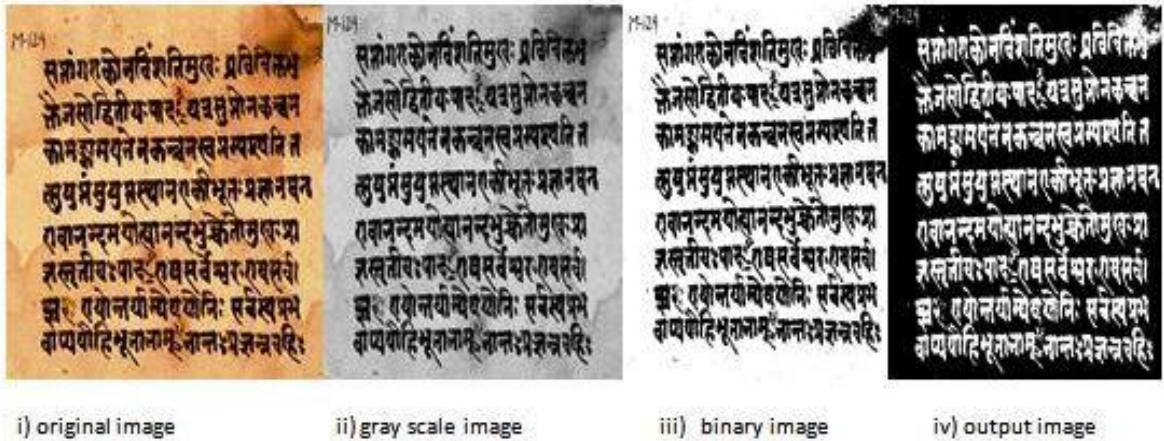

**Fig.2 Stages of the proposed method on document images**

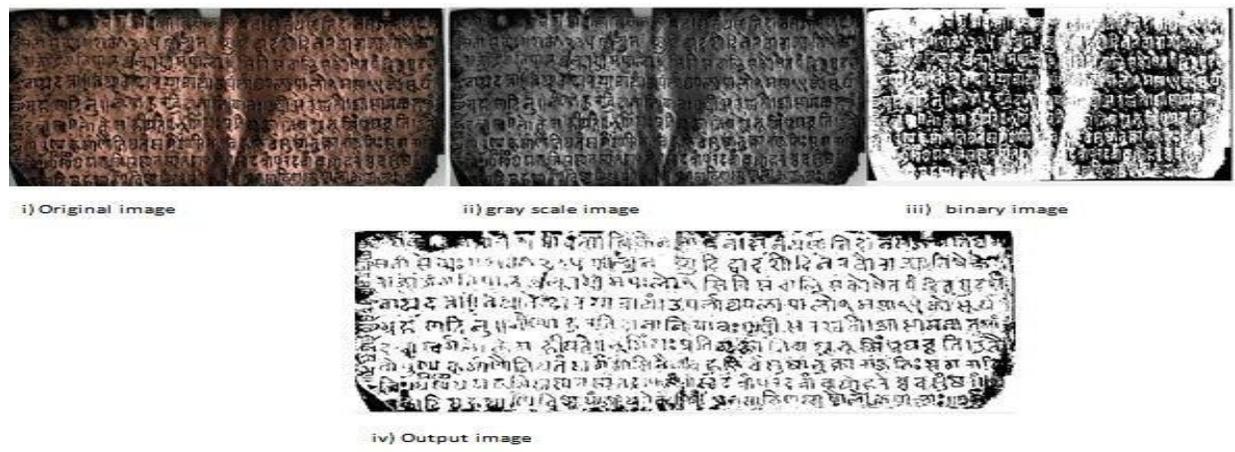

**Fig.3 Stages of the proposed method on metal plate images**

The first stage of the proposed method is binarization, which converts the input images into binary form. In a binary image, pixel values are represented using two levels, typically 0 and 1, corresponding to the background and foreground. Initially, the input RGB (Red, Green, Blue) image is converted into a grayscale image. The grayscale image is then binarized using an appropriate threshold value to separate text from the background [19].

Following binarization, suitable morphological operations are applied to refine the text regions and remove residual noise. The enhanced images are subsequently evaluated using K-Nearest Neighbor (K-NN) and Support Vector Machine (SVM) classifiers, with mean and standard deviation of local pixel neighborhoods used as feature descriptors. This approach effectively addresses low-contrast conditions in ancient script images, resulting in a cleaner background and improved readability [18], [20].

**Algorithm: Enhancement and Classification of Ancient Marathi Script Images**

**Input:** RGB images of ancient Marathi scripts
**Output:** Enhanced binary images and classification results

**Steps:**
1. **Image Acquisition**

    - Collect ancient Marathi script images from various sources (stone, metal plates, and documents).
    - Split dataset into training (80%) and testing (20%) sets.

2. **Preprocessing**
    a. Convert input RGB images to grayscale.
    b. Apply binarization to convert grayscale images into binary images using appropriate thresholding.

    - Use global thresholding for images with uniform background.
    - Use local thresholding for images with irregular background based on local mean and standard deviation.

3. **Morphological Operations**

    - Apply morphological techniques to refine text regions and remove residual noise.

4. **Feature Extraction**

    - Compute local mean μ and standard deviation σ for each text region:

    $$\mu = \frac{1}{n}\sum_{i=1}^{n} x_i$$

    $$\sigma = \sqrt{\frac{1}{n}\sum_{i=1}^{n}(x_i - \mu)}$$

    - These features describe local intensity variations and texture characteristics.

5. **Classification**

    - Train K-Nearest Neighbor (K-NN) and Support Vector Machine (SVM) classifiers using extracted features from training images.
    - Test the classifiers on the testing dataset to evaluate performance.

6. **Evaluation**

    - Measure classification accuracy for each image class (stone, metal plate, document).

- Compare results with existing methods in terms of accuracy and processing speed.

7. **Output**
    - Enhanced binary images with clean background.
    - Classification results indicating the type of script and readability improvement.

**Experimental Results**

For the experimental evaluation, a total of 250 images were used, consisting of images with both regular and irregular backgrounds. These images were collected from publicly available sources on the internet. The experiments were conducted using MATLAB R2016a. The images were tested using K-Nearest Neighbor (K-NN) and Support Vector Machine (SVM) classifiers, with features extracted from the mean and standard deviation of local pixel neighborhoods.

**Table 1: Distribution of Image Classes**

| Class Name | Background Type | No. of Images |
| --- | --- | --- |
| Stone Script | Normal (Regular) | 40 |
| Stone Script | Abnormal (Irregular) | 60 |
| Metal Script | Normal (Regular) | 20 |
| Metal Script | Abnormal (Irregular) | 80 |
| Document Script | Normal (Regular) | 30 |
| Document Script | Abnormal (Irregular) | 20 |
| **Total** | | **250** |

**Table 2: Sample images for various classes**

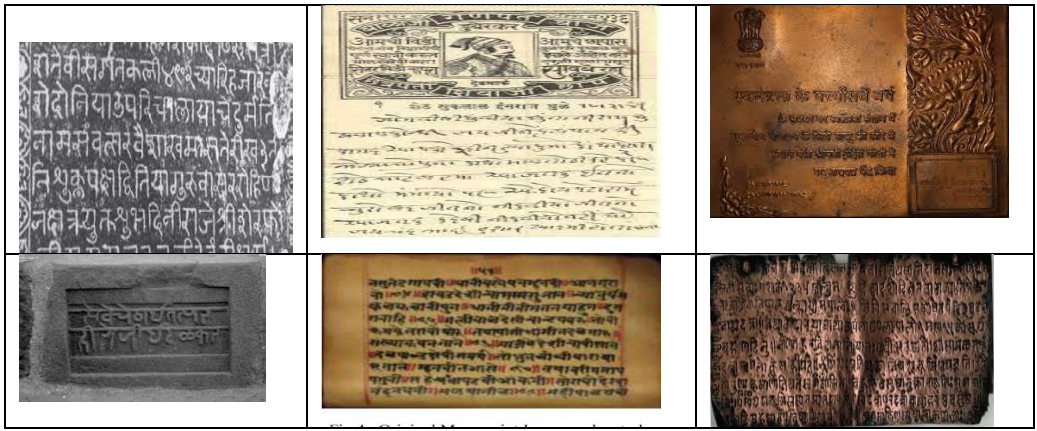

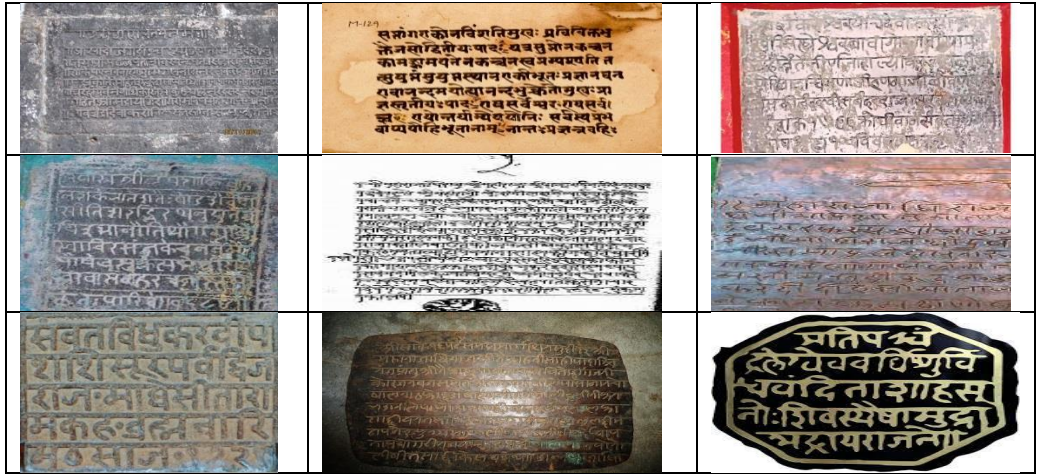

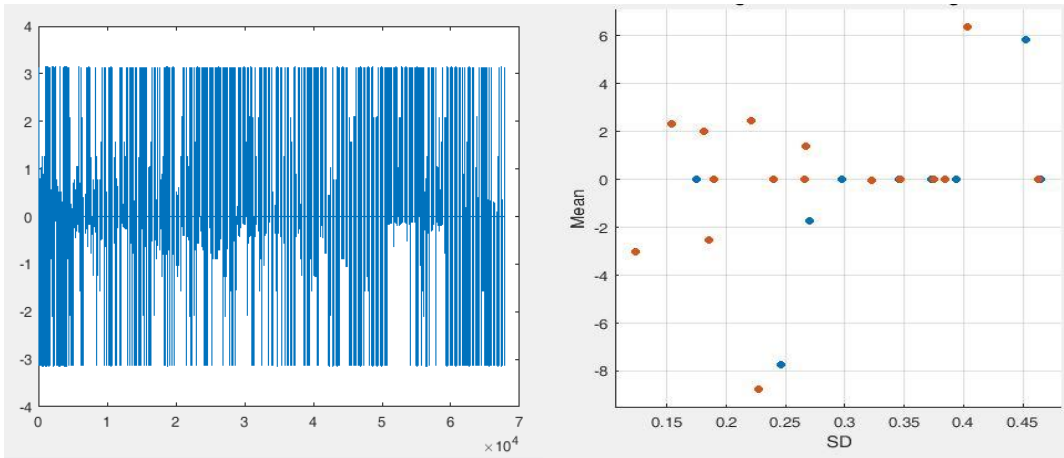

Fig. 4) Scatter plot showing classification results orange dots represents irregular background images and blue dots represent regular background images.

This dataset contains a diverse set of ancient Marathi script images with varying levels of contrast, background noise, and degradation. The proposed preprocessing and enhancement method is evaluated across all classes, and the performance is measured using the K-NN and SVM classifiers.

For experimental evaluation, 80% of the images were used for training, while the remaining 20% were reserved for testing. The training and testing sets were randomly selected to ensure balanced representation across all image classes. Figure 4 illustrates the frequency distribution of the extracted statistical features (mean and standard deviation), while Figure 5 presents a scatter plot showing the distribution of different image classes based on these features. These visualizations help to analyze feature separability and the effectiveness of the proposed preprocessing method in distinguishing between regular and irregular background images.

## Conclusion

In this paper, an image enhancement method for ancient Marathi script images with degraded backgrounds is presented, based on local mean and standard deviation statistics. Experimental results demonstrate that the proposed method achieves better performance on document-type images, with a maximum classification accuracy of 67.8% using the K-Nearest Neighbor (K-NN) classifier, compared to stone and metal plate scripts. The method has been successfully evaluated using both K-NN and Support Vector Machine (SVM) classifiers. Furthermore, when compared with existing techniques, the proposed approach shows improvements in both accuracy and processing speed, indicating its effectiveness for enhancing readability and supporting automated analysis of degraded ancient scripts.